\documentclass[conference]{IEEEtran}
\IEEEoverridecommandlockouts
\usepackage{amsmath,graphicx}
\usepackage{multirow}
\usepackage{adjustbox}
\usepackage{xcolor}
\usepackage[skins]{tcolorbox}
\usepackage{cite}
\usepackage{url}
\usepackage{booktabs}

\begin{document}
\title{Guiding ChatGPT to Generate Salient Domain Summaries}

\author{Jun Gao, Ziqiang Cao$^\ast$\thanks{$\ast$ Corresponding Author.}, Shaoyao Huang, Luozheng Qin, Chunhui Ai\\
School of Computer Science and Technology,
Institute of Artificial Intelligence, Soochow University\\
\{jgao1106, 20224227022, 20225227060, 20215227120\}@stu.suda.edu.cn; zqcao@suda.edu.cn}
\maketitle
\begin{abstract}
ChatGPT is instruct-tuned to generate general and human-expected content to align with human preference through Reinforcement Learning from Human Feedback (RLHF), meanwhile resulting in generated responses not salient enough.
Therefore, in this case, ChatGPT may fail to satisfy domain requirements in zero-shot settings, leading to poor ROUGE scores.
Inspired by the In-Context Learning (ICL) and retelling ability of ChatGPT, this paper proposes PADS, a \textbf{P}ipeline for \textbf{A}ssisting ChatGPT in \textbf{D}omain \textbf{S}ummarization.
PADS consists of a retriever to retrieve similar examples from corpora and a rank model to rerank the multiple candidate summaries generated by ChatGPT.
Specifically, given an inference document, we first retrieve an in-context demonstration via the retriever.
Then, we require ChatGPT to generate $k$ candidate summaries for the inference document at a time under the guidance of the retrieved demonstration.
Finally, the rank model independently scores the $k$ candidate summaries according to their quality and selects the optimal one.
We extensively explore dense and sparse retrieval methods to select effective demonstrations for reference and efficiently train the rank model to reflect the quality of candidate summaries for each given summarized document.
Additionally, PADS contains merely 400M trainable parameters originating from the rank model and we merely collect 2.5k data to train it.
We evaluate PADS on five datasets from different domains, and the result indicates that each module in PADS is committed to effectively guiding ChatGPT to generate salient summaries fitting different domain requirements.
Specifically, in the popular summarization dataset Gigaword, PADS achieves over +8 gain on ROUGE-L, compared with the naive ChatGPT in the zero-shot setting.
\footnote{Our code are available at \url{https://github.com/jungao1106/PADS}}
\end{abstract}

\section{Introduction}
The increasing volume of text from the web entails the development of summarization task \cite{cao2018faithful,wan2015multi}, aiming to generate concise and informative summaries while retaining the critical information of the original text.
Additionally, in the era of information explosion, summarization involves various domains, including news reports, scientific articles, and social media.
Therefore, a general cross-domain generation system is necessary in practical scenarios.

Recently, Large Language Models (LLMs) such as ChatGPT exhibited powerful emergence phenomenon through In-Context Learning (ICL) \cite{brown2020language,min2022rethinking,xie2021explanation} and Chain of Thought (CoT) \cite{wei2022chain}.
Therefore, researchers attempt to employ ChatGPT in text summarization~\cite{wang2023chatgpt,yang2023exploring} to get rid of fine-tuning models on specific datasets independently.
However, they point out that the summaries generated by ChatGPT outperform supervised models in human evaluation, while achieving poor ROUGE \cite{lin2004rouge} scores.
We attribute this to each domain having unique characteristics and requirements and ChatGPT is not able to always generate summaries fitting requirements of various domains without any guidance.
For instance, news summaries need to briefly introduce the entire events, while dialogue summaries only focus on the main part of conversations.
Specifically, ChatGPT is instruct-tuned to align with human preference with a trade-off for best performance, namely, the ``alignment tax''\footnote{The base model of ChatGPT is text-davinci-003, which is the supervised RLHF instructed tuned version of code-davinci-002. In many papers, researchers proved that the code-davinci-002 achieves the best performance but sometimes generates human unexpected contents.} proposed by OpenAI \cite{ouyang2022training}.
Therefore, the ``alignment tax'' makes ChatGPT generate more general and safe but less salient summaries, resulting in poor ROUGE scores.

This paper proposes PADS, a \textbf{P}ipeline for \textbf{A}ssisting ChatGPT in \textbf{D}omain \textbf{S}ummarization, to guide ChatGPT in performing domain summarization tasks based on in-context learning and retelling ability of ChatGPT.
In PADS, we first utilize a dense retriever, Sentence-BERT (S-BERT) \cite{reimers2019sentence}, to retrieve the most similar document based on cosine similarity from the corpora as in-context demonstrations.
Then, we provide the demonstration to ChatGPT combined with the inference document through multi-turn conversations and require ChatGPT to generate $k$ candidate summaries at a time.
Finally, the rank model scores the candidates and selects the optimal summary among them.
The dense retriever is initialized from previous officially released models~\footnote{https://huggingface.co/sentence-transformers/all-distilroberta-v1.} and the ChatGPT is employed on Azure called through web services, neither of which require training, hence, the merely trainable parameters originate from the rank model.
Specifically, the rank model is composed of an auto-regressive model and a scoring head, containing 400M trainable parameters.
We train the auto-regressive model by contrast learning to avoid feature collapse and to train the scoring head supervised by normalized rouge scores with 2.5k data for each dataset respectively.

We evaluate PADS on five datasets from different domains, and the results indicate that ChatGPT achieves performance gains in all datasets with the assistance of PADS.
Especially in the popular dataset Gigaword, PADS achieves over +8 gain on ROUGE-L, compared with the naive ChatGPT in the zero-shot setting.

The main contributions of this paper are as follows: 
\begin{itemize}
    \item We propose PADS, a parameter-efficient pipeline to guide ChatGPT in being competent in domain summarization. 
    \item We extensively explore sparse and dense retrieval methods for retrieving similar documents as powerful in-context demonstrations.
    \item We integrate contrastive learning in the rank model to measure summaries quality conditioned on each given document.
\end{itemize}

\section{Related Work}
\subsection{In-Context Learning}
In recent times, a lot of LLMs, including ChatGPT \cite{luo2023chatgpt}, GPT-4 \cite{openai2023gpt4}, LLaMA \cite{meta2023introducing}, and Chat-GLM \cite{du2022glm}, exhibit general spectacular emergent abilities that provides a novel paradigm for generative models known as Pre-training, Prompting, and Prediction.
Within this paradigm, in-context learning assumes a pivotal role, bolstering the generalization capability of LLMs \cite{goyal2022news,wang2023chatgpt,yang2023exploring,wei2022chain}, without necessitating gradient updating.

Presently, one research trajectory within the realm of in-context learning focuses on the retrieval of demonstrations for LLMs.
In its early stages, researchers employed random selection from the training dataset to obtain task-relevant examples as in-context demonstrations \cite{brown2020language}.
Subsequently, the few-shot capabilities of GPT-3 became associated with the choice of in-context examples \cite{liu2021makes}, and the sequence of in-context examples is now taken into deliberation \cite{lu2021fantastically} as well.
In this paper, we follow this research line and activate the pre-training knowledge of ChatGPT by providing well-retrieved demonstrations.

\subsection{Dense Retriever}
Dense retrievers~\cite{reimers2019sentence,wang2022text,wang2023learning} distinguish themselves as a prominent information retrieval methodology, utilizing dense vectors to expertly perform semantic alignment between queries and documents within latent space.
In sharp contrast to conventional sparse retrieval techniques such as BM25~\cite{robertson2009probabilistic}, dense retrieval capitalizes on the impressive modeling ability embedded within pre-trained language models (PLMs)~\cite{devlin2018bert}.
To further enhance the effectiveness of dense retrieval, a diverse array of strategies has surfaced.
These encompass techniques such as hard negative mining~\cite{karpukhin2020dense}, knowledge distillation~\cite{ren2021rocketqav2}, and continual pre-training~\cite{wang2023learning}, collectively amplifying the performance of dense retrieval.
In this paper, we extensively explore both sparse and dense retrieval methods and employ S-BERT as our retriever from their officially released models.

\subsection{Domain Summarization}
In the summarization task, the majority of the work focused on specific domains, such as science articles \cite{yasunaga2019scisummnet}, dialogues \cite{gliwa2019samsum}, and emails \cite{zhang2019email}.
Meanwhile, there have been several works exploring the concepts of domains\cite{hua2017pilot,gehrmann2018bottom}, such as associating domain-related knowledge with template information~\cite{cheung2013towards} and exploring domain shift in text summarization~\cite{wang2019exploring}.
Moreover, AdaptSum explored the domain and task adaptive pre-training and highlighted the challenges in domain summarization tasks \cite{yu2021adaptsum}.
These works relied on supervised methods, and they lacked generalization capability.
With the emergence of LLMs, zero-shot summarization experiments were performed on GPT-3 in the news domain and the results pointed out that the summaries generated by LLMs achieved low ROUGE scores \cite{goyal2022news}.
In PADS, we address the domain adaption problems by employing ChatGPT as the backbone and improve LLMs performance on automatic metrics by providing well-retrieved demonstrations and reranking candidate summaries.

\section{Problem Formulation}
Recently, researchers have highlighted that LLMs parameterize massive world knowledge during pre-training and viewed LLMs as lossless compressors~\cite{deletang2023language}.
In this paper, we delve into the realm of domain summarization knowledge of ChatGPT, viewing it as a subset of the broader spectrum of world knowledge.
Our objective is to harness and activate this knowledge through in-context learning.
To accomplish this, we build upon the foundation laid by previous in-context learning theoretical analyses~\cite{xie2021explanation}.
They revolve around the idea of Language Models (LMs) generating sequences by sampling tokens from a Hidden Markov Model (HMM)~\cite{baum1966statistical}.
We further extend this idea to the specific domain summarization task.
In the context of a given domain set represented as $d=(d_1, d_2, ..., d_n)$, our assumption is that ChatGPT precisely aligns with the pretraining distribution $P_\Theta$, encompassing the extent of world knowledge.
Consequently, the task of generating summaries that align with domain requirements boils down to the conditional distribution of given inference documents $P_\Theta(y|x)$ within the pretraining distribution.
To be more precise, this conditional distribution is a posterior predictive distribution achieved through marginalizing the elements within the domain set $d$, where $\sum_{d_i\in d} P_\Theta(d_i|x) = 1$:
\begin{equation}
    P_\Theta(y|x)=\sum_{d_i\in d} P_\Theta(y|d_i, x)P_\Theta(d_i|x).
\label{equ:icl}
\end{equation}
Naturally, the generated summaries are governed by the conditional distribution described in Equation \ref{equ:icl}, where $P_\Theta(d_i|x)$ plays a pivotal role in determining the relative importance of various domains account for.
In a zero-shot scenario, when the input consists solely of inference documents from the domain $d_*$, it is plausible that the output distribution allocates less attention to $d_*$, resulting in a more generalized yet less domain-specific output distribution denoted as $P_\Theta^-(y|x)$.
This corresponds to the normal situation where generated summaries fall short of meeting domain-specific requirements.

However, PADS has the potential to compel ChatGPT to attend more to the target domain $d_*$.
Specifically, it implicitly increases $P_\Theta(d_*|x)$ through the introduction of well-retrieved demonstrations into the input data. Consequently, we obtain a domain-specific output distribution $P_\Theta^+(d_*|x)$ after marginalization.
In our preliminary exploration, ChatGPT is able to generate multiple summaries at a time, denoted as $(y_1, y_2, ..., y_k)$.
These candidate summaries are subject to the domain salient output distribution $P_\Theta^*(y|x)$ that each summary $y_i$ is drawn from $P_\Theta^+(y|x)$, where $i$ ranges from 1 to $k$.
Importantly, the salient output distribution inherently offers a quality metric for each candidate summary, enabling us to rank these candidates based on $P_\Theta^+(y_i|x)$, where $i$ ranges from 1 to $k$.
Given that ChatGPT operates as a black-box model, we leverage the generation order of candidate summaries as a superficial representation of this metric.
In this context, the first candidate summary is considered the best one according to ChatGPT's judgment.
However, as this quality metric is provided by the generated model itself, we introduce a rank model to provide a supervised and more convincing quality metric denoted as $\mathcal{M}(y|x)$, aiming to further enhance the summary quality.
The values of $\mathcal{M}(y_i|x)$, with $i$ ranging from 1 to $k$, can be interpreted as the probabilities indicating the likelihood that the current candidate summary, $y_i$, is the best choice.
Finally, PADS selects the one with the highest probability.

\section{Methodology}
\begin{figure*}
    \centering
    \includegraphics[width=1\textwidth]{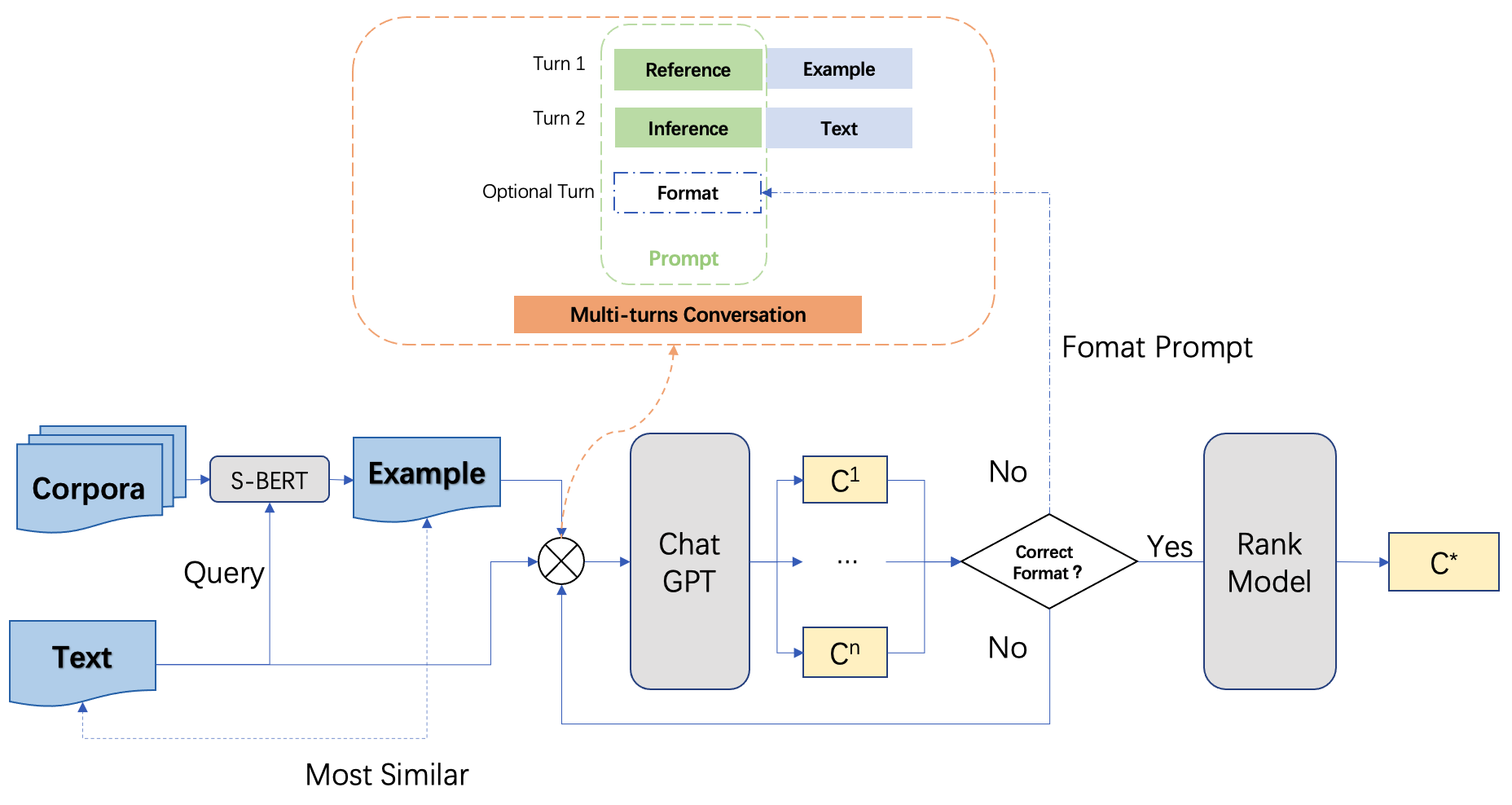}
    \caption{Workflow of PADS. Text is the user input to be summarized and corpora is the demonstrations to be retrieved. Reference and Inference are manually designed prompts in different conversation turns to distinguish inference documents from demonstrations. The optional format prompt is engaged in another conversation turns to correct the output format of ChatGPT if necessary.}
    \label{fig:fig1}
\end{figure*}

The workflow of PADS is shown in Figure \ref{fig:fig1}.
Firstly, given a text $T$ to be summarized, PADS retrieves similar demonstrations from the corpora, the training sets of each dataset in our experiment settings.
Then, we feed demonstrations combined with the inference document to ChatGPT in multi-turn form and require ChatGPT to generate multiple candidate summaries $(C^1, C^2,.., C^n)$ at a time, as demonstrated in Table~\ref{tab:template}.
Finally, the rank model scores the candidate summaries conditioned on the given documents independently to get the optimal summary $C^*$.

\subsection{Examples Retrieval}
\label{methodlogy:sec1}
Previous research results \cite{min2022rethinking} have shown that in-context learning with $n$ demonstrations $(x_1,y_1), (x_2,y_2), ..., (x_n,y_n)$ has minimal impact on model performance regardless of whether $y_i$ is correct or not.
Therefore, following the previous conclusion, we retrieve similar demonstrations based on the original text to be summarized, and we employ the dense retriever S-BERT to output latent features.
S-BERT can be regarded as an encoder $E(\cdot)$ and we measure the cosine similarity between latent features of inference documents $q$ and retrieval documents $k$:
\begin{equation}
    sim(q,k)=Cosine(E(q),E(k)).
\end{equation}

\subsection{Context Construction}
\begin{table}
\caption{The reference and inference prompts used to require ChatGPT generates $k$ candidates at a time under the guidance of provided demonstrations}
\begin{tcolorbox}

[USER] 

I will present you the text and its standard summary, considering it as an example. Text: \textcolor[rgb]{0,0,0.9}{\{Similar Document\}} Summary: \textcolor[rgb]{0,0,0.9}{\{Golden Summary\}}

[ASSISTANT]

...

[USER] 

Combining the above example, generate $k$ different summaries of the following text. Text: \textcolor[rgb]{0,0,0.9}{\{Inference Document\}} Summary: 

[ASSISTANT]\\
1. \textcolor[rgb]{0.9,0,0}{\{Candicate 1\}}\\
...\\
$k$. \textcolor[rgb]{0.9,0,0}{\{Candicate $k$\}}
\end{tcolorbox}

\label{tab:template}
\end{table}
It is well known that modeling historical conversation presented by supporting multi-turn conversation is one of the important natural abilities of ChatGPT as an outstanding chatbot, and therefore, instead of directly concatenating inference documents and demonstrations \cite{yang2023exploring,goyal2022news,wang2023cross}, we provide in-context demonstrations and format requirement in the multi-turn form, as shown in the orange box of Figure \ref{fig:fig1}.
Specifically, the multi-turn conversations are demonstrated in Table \ref{tab:template}, and we ignore the first round of responses from ChatGPT since we provide demonstrations for ChatGPT to reference in this round.
However, as a chatbot, the generated summaries may not always satisfy our format, which increases the complexity to post-processing.
Hence, we engage in another round of conversation with ChatGPT to emphasize the format requirements as the following instruction: \textit{Answer in this format: 1: xxx\textbackslash{n}...5: xxx} if the generated summaries fail to meet the format requirement.

\label{experiment:sec6}

\subsection{Summary Rerank}
\label{methodology:sec3}
\begin{figure}
    \centering
    \includegraphics[width=0.5\textwidth]{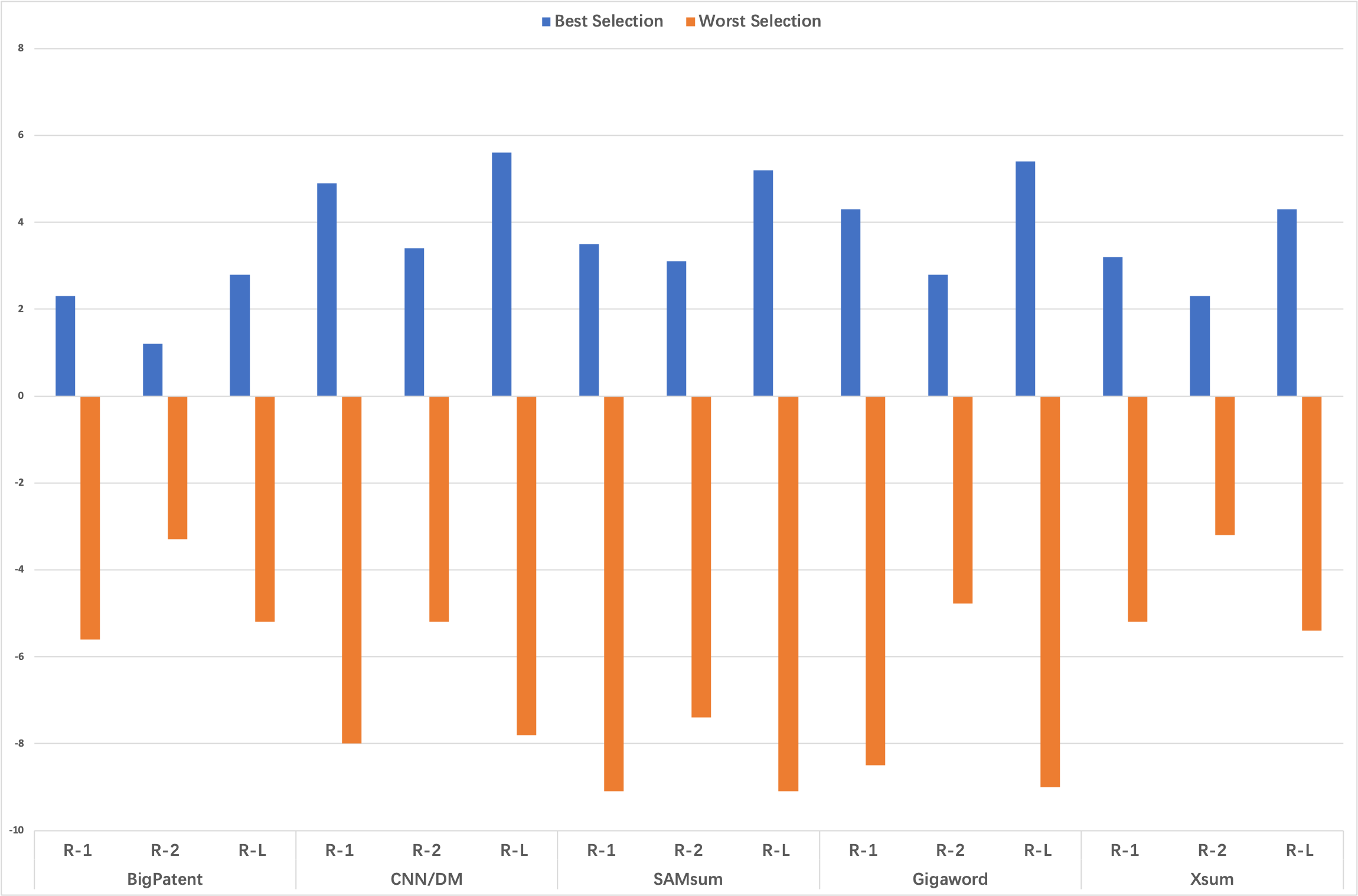}
    \caption{The relative ROUGE scores difference of summaries with highest and lowest scores among five candidate summaries. The first candidate's scores serve as the baseline.}
    % \end{adjustbox}
    \label{fig:fig3}
\end{figure}

In the preliminary experiments, we found that the summaries generated by ChatGPT were unstable, as shown in Figure \ref{fig:fig3}.
We assume the first summary is the best one that ChatGPT has the most confidence.
The significant relative differences indicate two problems.
On the one hand, the first summary that ChatGPT considers the best one is not the really best one; on the other hand, the performance of candidate summaries is unstable as well.
Therefore, the performance gap motivates us to train a summary rank model to select the potential optimal summary.
We adopt the large version of pre-trained Bart \cite{lewis2019bart} for its Encoder-Decoder architecture to rerank different candidate summaries.
As for the selection of the Encoder-Decoder model, there are two main reasons.
Firstly, in the text summarization task, the document is obviously much longer than the summary.
Therefore, if the rank model is Encoder-only, concatenating the source text and candidate summaries will be faced with terrible truncation intuitively.
More importantly, due to the current LMs being mostly transformer-based~\cite{vaswani2017attention}, the output continuations of the encoder-only model will mostly attend to the same document part rather than the candidate summaries due to self-attention~\cite{vaswani2017attention}, naturally leading to embedding collapse.
While the encoder and the decoder of Bart don't share parameters, the decoder communicates with the encoder output continuations to attend to the input documents through the cross-attention mechanism.
Therefore, Bart alleviates the embedding collapse problem to some extent and significantly extends the input window size.

\subsection{Training}
The trainable parameters originate from the rank model, and we process the training set by requiring ChatGPT to generate $k$ candidate summaries at a time.
Then, we normalize ROUGE scores between candidates and the golden summary as labels to train the scoring head.
What's more, we conduct the training process in two phases for the Bart model and its scoring head.
In the first phase, we perform average pooling on the continuations output by the decoder of Bart as the candidate summary feature, and the rank model is trained to distinguish candidate summaries of different qualities at the embedding level through contrastive learning.
Specifically, we consider the golden summaries as the anchor $y^*$, and the candidate with the highest ROUGE-L score is treated as the positive example $y^+$, while the remaining summaries are negative examples $y^-$.
We optimize the backbone Bart by InfoNCE loss~\cite{he2020momentum} to minimize the distance between $y^*$ and $y^+$ while maximizing the distance between $y^*$ and $y^-$, and $\tau$ is the temperature parameter setting to 0.8:
\begin{equation}
    L = -log\frac{exp(y^*\cdot y^+/\tau)}{\sum_{i=1}^k exp(y^*\cdot y_i/\tau)}.
\end{equation}
After contrastive learning, Bart distinguishes the candidate's quality at the continuation level.
Therefore, in the second phase, we design a regression task to train the scoring head optimized by Cross-Entropy loss under the supervision of normalized ROUGE-L scores, where the entire backbone is frozen.

\section{Experiments}
This section introduces experiments conducted on PADS.
We evaluate PADS in five summarization datasets from different domains.

\subsection{Dataset}
\label{experiment:sec1}

\begin{table}
\centering
\caption{An overview of datasets used in our experiments. 
Text/Summary Avg. tokens are the number of average tokens in the text/summary of test sets.
Train/Test is the size of the training/test set composing the original size/the quantity we sample.}
\begin{adjustbox}{width={0.5\textwidth},totalheight={\textheight},keepaspectratio}
\begin{tabular}{l|c|c|c|c|c}

\hline
      & BigPatent & CNN/DM & SAMsum  & Gigaword & XSum\\
\hline
Domain & Technology & News  & Social Media  & News & News  \\
\hline
Train & 1,207,222/2,500 & 287,113/2,499 & 14,732/2,500 & 3,803,957/2,500 & 204,045/2,500 \\
\hline
Test  & 67,072/1,481 & 11,490/1,500 & 819/791 & 1,951/1,500 & 11,334/1,479 \\
\hline
Text Avg. tokens & 3,574   & 691   & 93    & 31    & 358 \\
\hline
Summary Avg. tokens & 116    & 51    & 20     & 8     & 21 \\

\hline
\end{tabular}%
\end{adjustbox}

\label{tab:dataset}%
\end{table}%

We conduct experiments on five datasets from different domains, as shown in Table \ref{tab:dataset}.
Notably, although CNN/DM \cite{nallapati2016abstractive}, XSum \cite{Narayan2018DontGM}, and Gigaword \cite{graff2003english,Rush_2015} are all datasets in the field of news, CNN/DM focuses on sports and entertainment, XSum concentrates on tech and culture, while Gigaword focuses on economics. 
We sample 2,500 instances from each training set to train our rank model and prepare the training set by requiring ChatGPT to generate multiple candidates with the prompt: \textit{Generate k different summaries of
the following text. Text: {Inference Document} Summary:}.

Additionally, we select the entire test set from SAMsum~\cite{gliwa2019samsum} and we sample over 1.5k samples for other datasets.
More details about the involved datasets are as follows:

\subsubsection{BigPatent}
BigPatent\footnote{We employ the default 2.1.2 version with all CPC codes at \url{https://huggingface.co/datasets/big_patent.}}~\cite{sharma2019bigpatent} is a comprehensive dataset comprising 1.3 million U.S. patent documents, each accompanied by human-written abstractive summaries.
These patents are systematically classified into nine distinct categories using Cooperative Patent Classification (CPC) codes, encompassing a wide range of fields, from essential human necessities to cutting-edge technology.

\subsubsection{CNN/DM}
The CNN/DailyMail (CNN/DM)\footnote{\url{https://huggingface.co/datasets/cnn_dailymail.}}~\cite{nallapati2016abstractive}is an extensive collection of English-language news articles, encompassing more than 300,000 unique pieces authored by journalists from CNN and the Daily Mail.
While its original purpose was geared towards machine reading, comprehension, and abstractive question answering, the current version of the dataset is versatile, supporting both extractive and abstractive summarization tasks.

\subsubsection{SAMsum}
The SAMSum\footnote{\url{https://huggingface.co/datasets/samsum.}}~\cite{gliwa2019samsum} includes thousands of messenger-style conversations with summaries.
Professional linguists created these conversations to mimic real-life chats, covering various topics and styles, from informal to formal, including slang and typos.
The dataset was then annotated with third-person summaries of the conversations' content.

\subsubsection{Gigaword}
Gigaword\footnote{\url{https://huggingface.co/datasets/gigaword.}}~\cite{graff2003english,Rush_2015} stands as a renowned English summarization dataset, widely recognized for its single-line input documents that are meticulously curated from a diverse array of news sources.

\subsubsection{XSum}
The Extreme Summarization (XSum)\footnote{\url{https://huggingface.co/datasets/EdinburghNLP/xsum.}} \cite{Narayan2018DontGM} dataset serves as a valuable resource for assessing the performance of abstractive single-document summarization systems.
It comprises articles sourced from BBC publications spanning the years 2010 to 2017, encompassing an extensive array of domains, including but not limited to News, Politics, Sports, Weather, Business, Technology, Science, Health, Family, Education, Entertainment, and Arts.

\subsection{Evaluation Metrics}
ROUGE \cite{lin2004rouge} is a widely adopted metric in many generative tasks that evaluate how similar the generated hypothesis is to the golden reference.
Therefore, ROUGE is used in our experiments to evaluate the quality of summaries and we report the F-1 scores of ROUGE-1, ROUGE-2, and ROUGE-L (abbreviated R-1, R-2, R-L in the following), and we employed the files2rouge \footnote{https://github.com/pltrdy/files2rouge.} library in practice.

\subsection{Baselines}
\subsubsection{Supervised Method}
We employ Bart-large from huggingface fine-tuned on BigPatent, CNN/DM, SAMsum, XSum, and Gigaword as supervised models trained on the entire training set, respectively.
Notably, supervised models provide only reference performances instead of direct comparison in this paper, and therefore we don't select the State-of-the-Art (SOTA) models on each dataset.

\subsubsection{ChatGPT Zero}
Correspond to the zero-shot setting that ChatGPT generates one summary conditioned on the inference document only.

\subsubsection{ChatGPT Random}
We first provide a randomly sampled demonstration composed of a document and its golden summary to ChatGPT, and then ChatGPT generates $k$ summary conditioned on the inference document in the one-shot setting, while we retain the first one.

\subsubsection{ChatGPT Random (without Summary)}
Corespond to a variant of providing the random demonstration that provides documents only.

\subsubsection{ChatGPT Similar}
We first provide a retrieved demonstration composed of a document and its golden summary to ChatGPT, and then ChatGPT generates $k$ summary conditioned on the inference document in the one-shot setting, while we retain the first one.

\subsubsection{ChatGPT Similar (without Summary)}
Corespond to a variant of providing retrieved demonstrations that provides documents only.

\subsection{Settings}
In PADS, we empirically set the number of candidate summaries $k$ to 5.
To train the rank model, we fix the learning rate to 1e-6 and 3e-4 to train the Bart-large and scoring head, respectively, employing Adam as the optimizer.
The backbone of ChatGPT is gpt-3.5-turbo, we complete our experiments from November 6, 2023, to December 21, 2023.
Additionally, we call the API of ChatGPT supported by Azure and conduct other experiments on 8*NVIDIA A5000 24G GPUs.

When processing training sets and further inference, we randomly sample 2,500 instances from each training set and 1,500 instances for each test set, fixing them as origin data.
Then we require ChatGPT to generate five candidates as mentioned above.
We give 5 chances to emphasize formation for each instance, and some instances are filtered by the OpenAI usage policy\footnote{https://openai.com/policies/usage-policies.}.
Therefore, the effective training and test data may be less than the original instances to keep the validity of PADS.
\subsection{Result}
\label{experiment:sec4}
\begin{table*}
\centering
\caption{The performance of PADS and baselines. SFT is the supervised Bart-large model. w/o summary stands for provide the document only.}
\begin{adjustbox}{width={0.95\textwidth}}
% Table generated by Excel2LaTeX from sheet 'Sheet13'
\begin{tabular}{l|ccc|ccc|ccc|ccc|ccc}
\toprule
\multirow{2}{*}{Method} & \multicolumn{3}{c|}{BigPatent} & \multicolumn{3}{c|}{CNN/DM} & \multicolumn{3}{c|}{SAMsum} & \multicolumn{3}{c|}{Gigaword} & \multicolumn{3}{c}{XSum} \\
& R-1   & R-2   & R-L   & R-1   & R-2   & R-L   & R-1   & R-2   & R-L   & R-1   & R-2   & R-L   & R-1   & R-2   & R-L \\
\midrule
SFT   & 32.5  & 13.1  & 28.2  & 36.5  & 16.5  & 33.8  & 52.3  & 28.1  & 48.4  & 35.1  & 17.2  & 32.8  & 43.8  & 21.7  & 36.1 \\
\midrule
ChatGPT Zero & 36.7  & 11.2  & 31.7  & 28.7  & 10.8  & 25.9  & 32.7  & 11.1  & 29.1  & 17.7  & 4.3   & 15.1  & 19.4  & 5.7   & 16.0 \\
\midrule
ChatGPT Random & 36.9  & 11.1  & 31.7  & 29.6  & 11.0    & 26.5  & 38.6  & 15.3  & 34.3  & 21.1  & 6.7   & 18.2  & 24.9  & 12.6  & 21.3 \\
w/o summary & 31.3  & 8.5   & 26.6  & 24.2  & 6.4   & 20.6  & 29.7  & 7.2   & 24.9  & 16.6  & 3.3   & 12.9  & 19.8  & 4.4   & 14.5 \\
\midrule
ChatGPT Similar & 38.5  & 13.2  & 33.3  & 29.3  & 10.9  & 26.3  & 38.6  & 15.4  & 34.5  & 25.0  & 8.7   & 22.2  & 21.1  & 6.9   & 17.2 \\
w/o summary & 36.3  & 11.0    & 31.4  & 28.9  & 10.7  & 26    & 35.3  & 12.9  & 31.0  & 19.2  & 5.2   & 16.3  & 19.9  & 5.9   & 16.2 \\
\midrule
\textbf{PADS (ours)} & 38.6  & 13.3  & 33.7  & 33.2  & 12.3  & 29.5  & 40.9  & 15.6  & 36.0  & 26.4  & 8.6   & 23.4  & 26.8  & 8.7   & 21.3 \\
\bottomrule
\end{tabular}%
\end{adjustbox}
\label{tab:main}%
\end{table*}%

We evaluate PADS on five datasets from different domains in Table \ref{tab:main}, and the result indicates that PADS achieves significant performance gains and both the retriever and the rank model are necessiary.
Compared with the ChatGPT Zero, we can conclude that the task-relevant examples, whether similar or not, can improve the performance of ChatGPT in ROUGE scores.
At the same time, when provide retrieved demonstrations, ChatGPT can generate more salient summaries compared with the random task-relevant ones.
Subsequently, in the text summarization task, we prove that performance degradation occurs if golden summaries are not provided in both ChatGPT Random and ChatGPT Similar.
Finally, compared with one-shot settings, PADS further achieves performance gains since the rank model effectively reflect the quality among candidate summaries and correctly scores 5 candidates and gets a more salient summary finally.

\subsection{Case Study}
\label{experiment:case}
\begin{table*}
\centering
\caption{Case study on XSum dataset.
\textcolor[rgb]{ 0,  0,  1}{Blue} represents the cause or origin, and \textcolor[rgb]{ 1,  0,  0}{red} stands for the outcome.}
\begin{tabular}{l|p{0.9\textwidth}}
\hline
Reference & \textcolor[rgb]{ 1,  0,  0}{The High Court has begun hearing the legal challenge of a terminally ill UK man who wants the right to die.} \\
\hline
ChatGPT Zero & \textcolor[rgb]{ 0,  0,  1} {Noel Conway, a 67-year-old man with motor neurone disease, is asking the High Court in London to allow a doctor to prescribe him a lethal dose when his health deteriorates further. Conway said he wants to say goodbye to loved ones ``at the right time, not to be in a zombie-like condition suffering both physically and psychologicall''.} \textcolor[rgb]{ 1,  0,  0} {Anyone who assists in his death could face up to 14 years in prison. This change to the law would only apply to terminally ill adults with less than six months to live and who have a settled wish to die. The case is expected to take up to four days.} \\
\hline
\textbf{PADs (ours)}  & \textcolor[rgb]{ 0,  0,  1}{A terminally ill man named Noel Conway has started a legal battle to have the right to die, calling for a change in the law which would allow doctors to prescribe lethal doses to some sick patients who want to end their lives.} \textcolor[rgb]{ 1,  0,  0} {His case will be heard through the High Court and is set to reopen the contentious public debate surrounding the issue.} \\
\hline
\end{tabular}%
\label{tab:casestudy}%
\end{table*}%
We demonstrate a case study on XSum to provide an intuitive comparison between PADS and zero-shot ChatGPT in Table \ref{tab:casestudy}. 
Empirically, news summaries should focus on the cause and result of the main events or topics covered in news articles and should be concise and informative.
In the zero-shot setting, the generated lengthy summary includes the entire background of the event, information about the individuals involved, and even dialogues, which is not suitable for a news summary.
The summaries output by PADS remove the uninformative contents but contain the outcome (blue color) and succinctly summarize the event's cause or origin (red color).
To some extent, the summary output by PADS is more aligned with the news summarization.
\section{Analysis}
\label{experiment:sec5}
\subsubsection{Comparison of Retrieval Methods}
\begin{figure}[ht]
  \centering
  \includegraphics[width=0.5\textwidth]{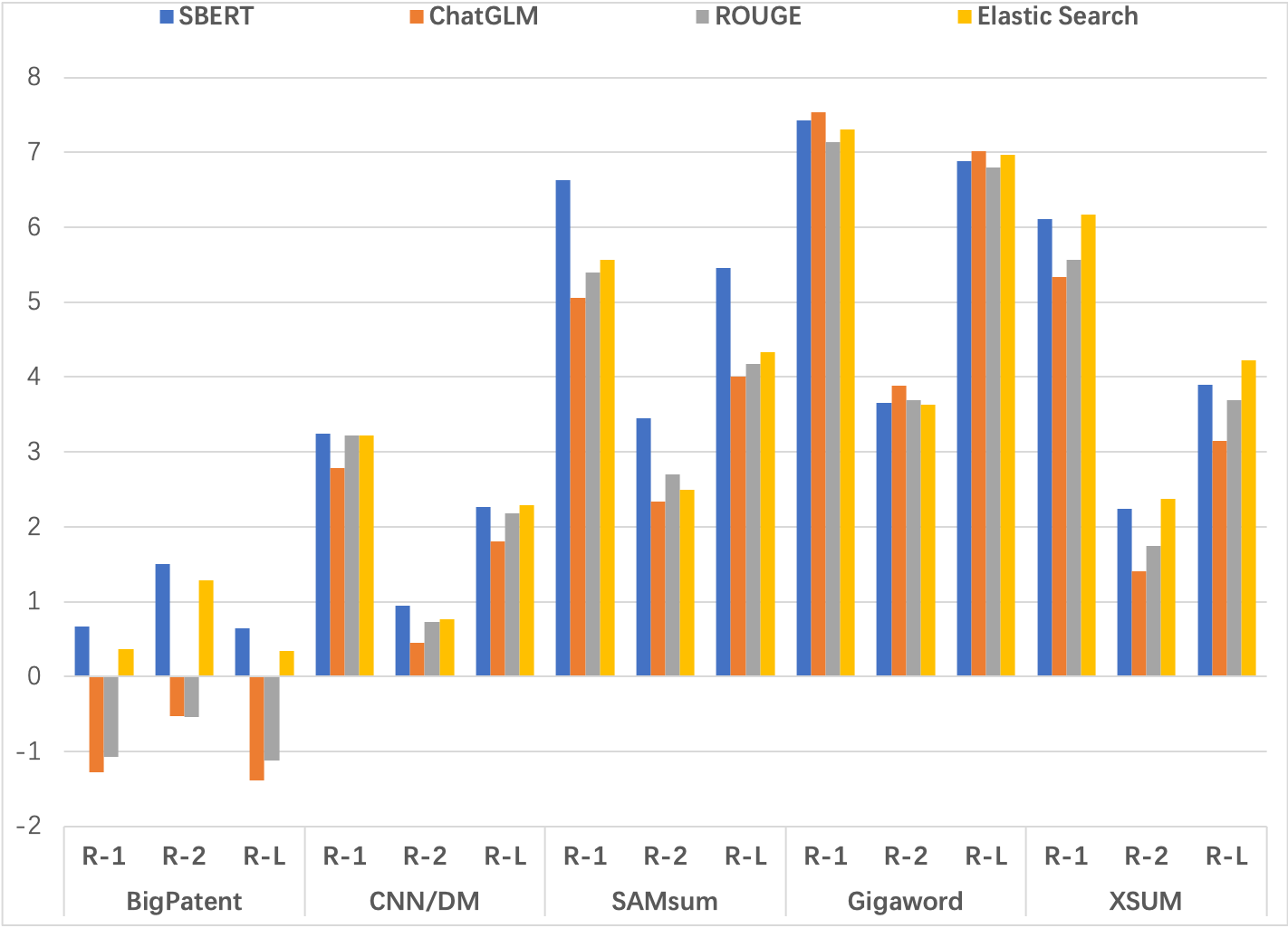}
  \caption{The relative ROUGE scores of ChatGPT Similar. The scores of ChatGPT Zero serve as the baseline.}
  \label{fig:retrievalmethods}
\end{figure}
To select a powerful retriever, we extensively conduct experiments on both sparse and dense retrievers in Figure~\ref{fig:retrievalmethods}, including S-BERT, ChatGLM-6b~\cite{du2022glm,zeng2023glm-130b}, ROUGE, and Elastic Search\footnote{\url{https://www.elastic.co}}, respectively, and the experiment data are 1,500 instances sampled from the validation set of each dataset, respectively.
Among them, ROUGE and Elastic Search are sparse retrievers, where Elastic Search is a popular retrieving engine based on BM25 algorithm~\cite{robertson2009probabilistic}, and S-BERT and ChatGLM-6b are employed as dense retrievers.
Specifically, the F-1 scores of R-L are used to measure similarity at the token level and we use the output continuations of the mask token in both ChatGLM and S-BERT to measure sequence similarity.
The result indicates that S-BERT outperforms the other three strategies in retrieving similar examples to guide ChatGPT to generate salient summaries.
Notably, although ChatGLM overwhelms S-BERT in model scale and hidden dimension and both have bi-direction encoder, its performance is worse than expected.
We attribute this to the pre-training task of the generative model, ChatGLM, naturally creates a gap in measuring sequence similarity compared with S-BERT without supervised tuning.

\subsubsection{Effectiveness of Multi-turn Conversation}
\begin{table}
\centering
\caption{The number of format-qualified instances generated in ChatGPT Similar. Concatenation represents directly concatenate examples and inference documents.}
\begin{adjustbox}{width={0.5\textwidth},totalheight={\textheight},keepaspectratio}
\begin{tabular}{c|c|c|c|c|c}
\hline
Methods & BigPatent  & CNN/DM  & SAMsum  & Gigaword & XSum  \\ \hline
Concatenation    & 2,070/2,500 & 2,004/2,500 & 2,459/2,500    & 2,348/2,500  & 2,145/2,500      \\ \hline
Multi-turn   & 2,500/2,500 & 2,499/2,500     & 2,500/2,500   & 2,500/2,500         & 2,500/2,500      \\ \hline
\end{tabular}
\end{adjustbox}

\label{tab:multiturn}
\end{table}

To construct training sets to train the rank model, we randomly sample 2,500 instances from each training set and require ChatGPT to generate 5 candidate summaries. 
Compared with direct concatenation, the number of format-qualified instances increases significantly when utilizing the multi-turn conversation ability of ChatGPT in Table \ref{tab:multiturn}.
Specifically, combined with Table \ref{tab:dataset}, we can conclude that as the documents are longer, such as BigPatent and CNN/DM, ChatGPT becomes confused to perform generation under specific requirements of format and quantity by concatenation.
However, longer documents seem no matter to multi-turn form since the outstanding language modeling ability of ChatGPT for historical dialogue as a chatbot.
\subsubsection{Theoretical Upper Bound}
The most similar demonstration is the inference example itself, and we consider this setting as the upper bound of ChatGPT Similar.
We first provide the inference document along with its golden summary to ChatGPT as demonstrations, and then, we require ChatGPT to generate one summary.
As shown in Table \ref{tab:upperbound}, PADS even outperforms the theoretical upper bound in XSum and Gigaword when provided with only well-retrieved demonstrations.
The result indicates that despite providing real answers as a reference, the first summary generated by ChatGPT may not always be the best one, further proving the necessity of the rank model in PADS.

\begin{table}
\centering
\caption{The theoretical upper bound in ChatGPT Similar.}
\begin{adjustbox}{width={0.5\textwidth},totalheight={\textheight},keepaspectratio}

% Table generated by Excel2LaTeX from sheet 'Sheet13'
\begin{tabular}{l|ccc|ccc|ccc|ccc|ccc}
\hline
\multirow{2}{*}{Method} & \multicolumn{3}{c|}{BigPatent} & \multicolumn{3}{c|}{CNN/DM} & \multicolumn{3}{c|}{SAMsum} & \multicolumn{3}{c|}{Gigaword} & \multicolumn{3}{c}{XSum} \\
& R-1   & R-2   & R-L   & R-1   & R-2   & R-L   & R-1   & R-2   & R-L   & R-1   & R-2   & R-L   & R-1   & R-2   & R-L \\
\hline
Upper Bound & 41.1  & 16.0  & 35.8  & 33.8  & 15.2  & 30.6  & 42.5  & 20.0  & 38.2  & 23.7  & 8.3   & 20.8  & 25.0  & 11.5  & 20.9 \\
\hline
\textbf{PADs (ours)} & 38.6  & 13.3  & 33.7  & 33.2  & 12.3  & 29.5  & 40.9  & 15.6  & 36.0  & 26.4  & 8.6   & 23.4  & 26.8  & 8.7   & 21.3 \\
\hline
\end{tabular}%

\end{adjustbox}
\label{tab:upperbound}%
\end{table}%

\section{Conclusion and Further Work}
This paper proposes PADS, composed of a retriever and rank model, to guide ChatGPT to generate salient domain summaries.
We extensively explore which retrieval strategy is better for retrieving demonstrations and how to make ChatGPT comprehend in-context demonstrations effectively.
Additionally, we prove that golden summaries of demonstrations are indispensable in summarization tasks during in-context learning.
Unfortunately, a latency analysis is necessary for PADS.
However, it's hard for us to effectively evaluate the inference latency brought by extra modules because we call the ChatGPT through the web service, although Microsoft Azure makes our experiment environment more stable than calling the API directly.
The retrieval cost is proportional to the corpus size and can be omitted through the offline preprocess, and the rerank cost can be ignored because the number of candidates is 5, and even a consumed GPU can load a BART to perform reranking in a second.
Therefore, ChatGPT generation time accounts for the majority of the entire Pipeline process, but this time is affected by the network and Azure load, and has a large fluctuation.
Finally, we believe there is much improvement room for PADS.
On the one hand, we plan to utilize the understanding ability of LLM to retrieve more powerful examples for guidance.
On the other hand, we only provide one demonstration in this paper because of maximum input limitation and charging for tokens, and we plan to compress the demonstrations in further work.
\newpage
\bibliographystyle{IEEEtran}
\bibliography{ref}

\end{document}